\documentclass[letterpaper]{article} 
\usepackage{aaai25}
\usepackage{times}  
\usepackage{helvet}  
\usepackage{courier}  
\usepackage[hyphens]{url}  
\usepackage{graphicx} 
\usepackage{natbib}  
\usepackage{caption} 
\usepackage{algorithm}
\usepackage{algpseudocode}

\usepackage{newfloat}
\usepackage{listings}

\usepackage{subfigure}
\usepackage{cite}
\usepackage{amsmath,amssymb,amsfonts}
\usepackage{amsthm}
\usepackage{booktabs}
\newtheorem{remark}{Remark}

\usepackage{tabularx}
\usepackage{array}
\newcolumntype{R}[1]{>{\raggedleft\let\newline\\\arraybackslash\hspace{0pt}}m{#1}}

\usepackage[dvipsnames]{xcolor}
\usepackage{tikz}  
\usepackage{bm}
\usepackage{pifont}
\usetikzlibrary{fit,calc}

\definecolor{mypink}{HTML}{FFEAEA}
\definecolor{myblue}{HTML}{DBF2F9}
\definecolor{tblblue}{HTML}{0801FF}
\definecolor{tblred}{HTML}{FF0000}
\newcommand{\boxit}[2]{
    \tikz[remember picture,overlay] \node (A) {};\ignorespaces
    \tikz[remember picture,overlay]{\node[yshift=4.75pt,fill=#1,opacity=1.0,fit={($(A)+(0,0.15\baselineskip)$)($(A)+(.99\linewidth,-{#2}\baselineskip - 0.25\baselineskip)$)}] {};}\ignorespaces
}

\DeclareCaptionStyle{ruled}{labelfont=normalfont,labelsep=colon,strut=off} 
\floatstyle{ruled}
\newfloat{listing}{tb}{lst}{}
\floatname{listing}{Listing}
\title{Asynchronous Federated Clustering with Unknown Number of Clusters \\ }
\author {
    Yunfan Zhang\textsuperscript{\rm 1},
    Yiqun Zhang\textsuperscript{\rm 1}\thanks{Coresponding Author},
    Yang Lu\textsuperscript{\rm 2},
    Mengke Li\textsuperscript{\rm 3},
    Xi Chen\textsuperscript{\rm 4},
    Yiu-ming Cheung\textsuperscript{\rm 4}
}
\affiliations {
    \textsuperscript{\rm 1}School of Computer Science and Technology, Guangdong University of Technology, Guangzhou, China\\
    \textsuperscript{\rm 2}Fujian Key Laboratory of Sensing and Computing for Smart City, School of Informatics, Xiamen University, Xiamen, China\\
    \textsuperscript{\rm 3}College of Computer Science and Software Engineering, Shenzhen University, Shenzhen, China\\
    \textsuperscript{\rm 4}Department of Computer Science, Hong Kong Baptist University, Hong Kong SAR, China\\
    \{yfzhang, yqzhang\}@gdut.edu.cn, luyang@xmu.edu.cn, csmengkeli@gmail.com, \{csxchen, ymc\}@comp.hkbu.edu.hk
}

\usepackage{bibentry}

\begin{document}

\maketitle

\begin{abstract}
Federated Clustering (FC) is crucial to mining knowledge from unlabeled non-Independent Identically Distributed (non-IID) data provided by multiple clients while preserving their privacy. Most existing attempts learn cluster distributions at local clients, and then securely pass the desensitized information to the server for aggregation. However, some tricky but common FC problems are still relatively unexplored, including the heterogeneity in terms of clients' communication capacity and the unknown number of proper clusters $k^*$. To further bridge the gap between FC and real application scenarios, this paper first shows that the clients' communication asynchrony and unknown $k^*$ are complex coupling problems, and then proposes an Asynchronous Federated Cluster Learning (AFCL) method accordingly. It spreads the excessive number of seed points to the clients as a learning medium and coordinates them across the clients to form a consensus. To alleviate the distribution imbalance cumulated due to the unforeseen asynchronous uploading from the heterogeneous clients, we also design a balancing mechanism for seeds updating. As a result, the seeds gradually adapt to each other to reveal a proper number of clusters. Extensive experiments demonstrate the efficacy of AFCL. 
\end{abstract}

%
\begin{links}
    \link{Code}{https://github.com/Yunfan-Zhang/AFCL}
\end{links}

\section{Introduction}\label{sec:intro}

Federated Learning (FL) is common in implementing distributed machine learning while preserving privacy \cite{review2022,Zhang2021,PrivacyFL-CIKN23}. In unsupervised FL tasks, Federated Clustering (FC) that partitions a dataset into compact object clusters demonstrates great potential in mining data concepts and knowledge \cite{Jing-CIKM19, Nelus2021, chung2022federated}. However, without label guidance, FC faces significant challenges brought by the privacy protection requirements and non-IID of heterogeneous clients. 

Most existing methods address FC by first letting clients learn cluster distributions and then passing the privacy-protected cluster knowledge to the server for global aggregation \cite{NEURIPS2020_e32cc80b, kumar2020federated}. For example, $k$-FED~\cite{k-fed} explores global cluster distributions with a higher security level through one-shot aggregation of the non-IID distributions learned by the clients. Two independent works, F-FCM \cite{F-FCM} and FFCM \cite{FFCM}, share similar names and principles, and adopt fuzzy-$c$-means as their local clustering algorithm. Since the fuzzy object-cluster affiliation can more finely reflect the partition information of data objects, the information loss caused by the privacy constraints of FL can be considerably offset. 

\begin{figure}[!t]
\centering
\includegraphics[width=0.44\textwidth]{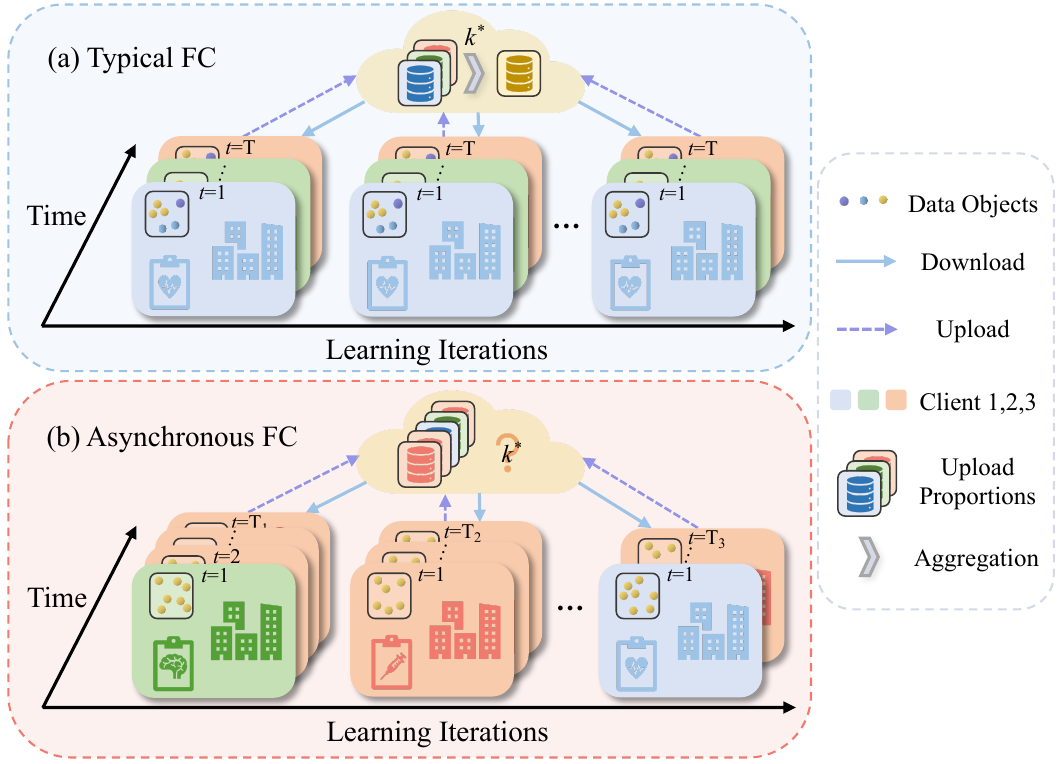}
\caption{\textbf{AFCL (ours) vs. typical FC.} (a) Existing FC approaches typically assume that the clients are synchronous and the optimal $k^*$ is known by both the clients and server. By contrast, (b) AFCL learns under a more realistic scenario that the clients can upload distribution information of completely non-overlapping and non-uniform numbers of clusters with unforeseen and imbalanced frequencies.}\label{fg:intro}
\end{figure}

Notably, most existing works overlook the common problem of client asynchrony\footnote{The terms ``asynchrony'' and ``asynchronous communication'' in this paper indicate the non-uniform participation rates of different clients in each round of communication. Since the goal of FC is to aggregate the distributions of clients at the server, the asynchronous problem addressed in this work is conceptually different from the model update asynchronous problem in supervised FL.} attributed to the divergence of clients' communication capabilities. Due to the lack of data labels, such a problem can severely bias the learning towards the clients that upload their distributions more frequently. Accordingly, a recently proposed federated spectral clustering method \cite{qiao2024federated} copes with the asynchronous scenario by aggregating intermediate variables to construct a global similarity matrix, which is robust to the communication frequency bias. However, it still relies on the naive assumption that the true number of clusters $k^*$ is known in advance, which limits the application domain of FC in many real applications with unforeseen $k^*$.

Automatic determination of the optimal $k^*$ has been an attractive clustering research topic in recent decades. Silhouette Coefficient \cite{Silhouettes1987} determines $k^*$ by considering the intra-compactness and the inter-dispersion of clusters. Density-based clustering \cite{schubert2017dbscan, zhangcais2024, peng2025icassp} performs cluster exploration by considering the distribution connection of objects, and can select $k^*$ according to the quality of cluster partition. More advanced learning-based approaches \cite{ahalt1990competitive, Yiu-mingCheung2005, jia2014cooperative, cai2024robust, zou2024federated} have been proposed to learn $k^*$ by letting seeds compete or cooperate to eliminate redundant low-quality clusters. Recently, significance-based clustering \cite{hu2022significance, hu2025significance} has been proposed using gap statistics to estimate $k^*$. Nevertheless, their learning process depends on detailed and sufficient data statistics, preventing them from being utilized in FC. Therefore, the absence of $k^*$ brings difficulties to FC due to the lack of clustering guidance, and also incurs additional $k^*$ learning objective, collectively making asynchronous FC a challenging task.

A more detailed comparison of asynchronous FC and typical FC are illustrated in Fig.~\ref{fg:intro}. Unlike typical FC where the clients communicate synchronously with known $k^*$, our focused asynchronous FC poses the problem of learning clusters and cluster numbers from the uploaded complex distributions caused by the non-uniform communication of clients. More specifically, the difficulties of asynchronous FC lie in the cross-coupled $k^*$ learning and asynchronous uploading of clients. That is, distributions learned at different clients are not associated with a uniform $k^*$, while the server struggles in the learning of $k^*$ due to the asynchronously uploaded unreliable local distributions. 

Therefore, we propose the Asynchronous Federated Cluster Learning (AFCL) method that can learn an optimal number of clusters with the asynchronously communicated clients in a self-adaptive way. It uniformly generates seed points for the clients, and accumulates the distribution information of their surrounding objects indicated by the difference between the seed and objects within each client to capture the clients' own distributions. Then the accumulated information is passed to the server to update the seeds for client-to-seed distribution information fusion. To gain a consensus among the clients, we let the neighboring seeds share their update intensity to achieve seed-to-seed information completion, as the non-IID clients may provide only a partial global distribution. A balancing mechanism has also been developed to evaluate and adjust the update intensity accumulated from different clients, to relieve the potential bias caused by their asynchronous participation. As a result, AFCL can automatically converge redundant neighboring seeds to learn an appropriate number of clusters under the challenging asynchronous federated scenario. Extensive experiments demonstrate the effectiveness of AFCL. The main contributions of this work are summarized into three points:
\begin{itemize}
    \item We propose a new FC approach to learn a distribution consensus from asynchronously communicated clients without requiring the `true' number of clusters. It serves to enhance the robustness and universality of FL.
    \item This paper first considers and attempts to solve a more realistic but challenging non-IID case in FC, i.e., a global cluster could be composed of completely non-overlapping sub-clusters belonging to different clients.
    \item A balancing mechanism is developed to allocate the contribution of clients with heterogeneous communication modes during the interactive learning on the server, even if they participate in the learning by only one-shot.
\end{itemize}

\section{Related Work}

\subsection{Federated Clustering}
An one-shot FC approach called $k$-FED \cite{k-fed} has been developed to relieve the leakage of information during communication, while FedKKM \cite{zhou2022memory} has been proposed to improve communication efficiency by using a novel Lanczos algorithm in its distributed matrices. Meanwhile, F-FCM \cite{F-FCM}, and FFCM \cite{FFCM} utilized fuzzy clustering techniques to enhance privacy protection by only transmitting object-cluster affiliation. Furthermore, A multi-view FC approach \cite{hu2023efficient} has been proposed to extend multi-view clustering into the federated scenario by designing a strategy of consensus prototype learning. However, they employ clustering techniques under the assumption that the true cluster numbers are known by the clients and server in advance. Recently, an FC framework VKMC \cite{huang2022coresets} has been proposed to improve the vertical FL based on coresets, while a density-based FC method HFDPC \cite{ding2023horizontal} has been proposed for improving the effectiveness of data partitioning by introducing a similar density chain. Most recently, Federated Subspace Clustering (Fed-SC) \cite{xie2023fed} and Federated Spectral Clustering (FedSC) \cite{qiao2024federated} have been proposed to address the FC of high-dimensional and noisy data, respectively. Although some of the above-mentioned FC methods have considered the non-IID or the asynchrony issues of clients in FC, most of their solutions heavily depend on the availability of the `true' number of clusters $k^*$, which hinders their applications in real complex scenarios.

\begin{figure*}[!t]
\centering
\includegraphics[width=0.98\linewidth]{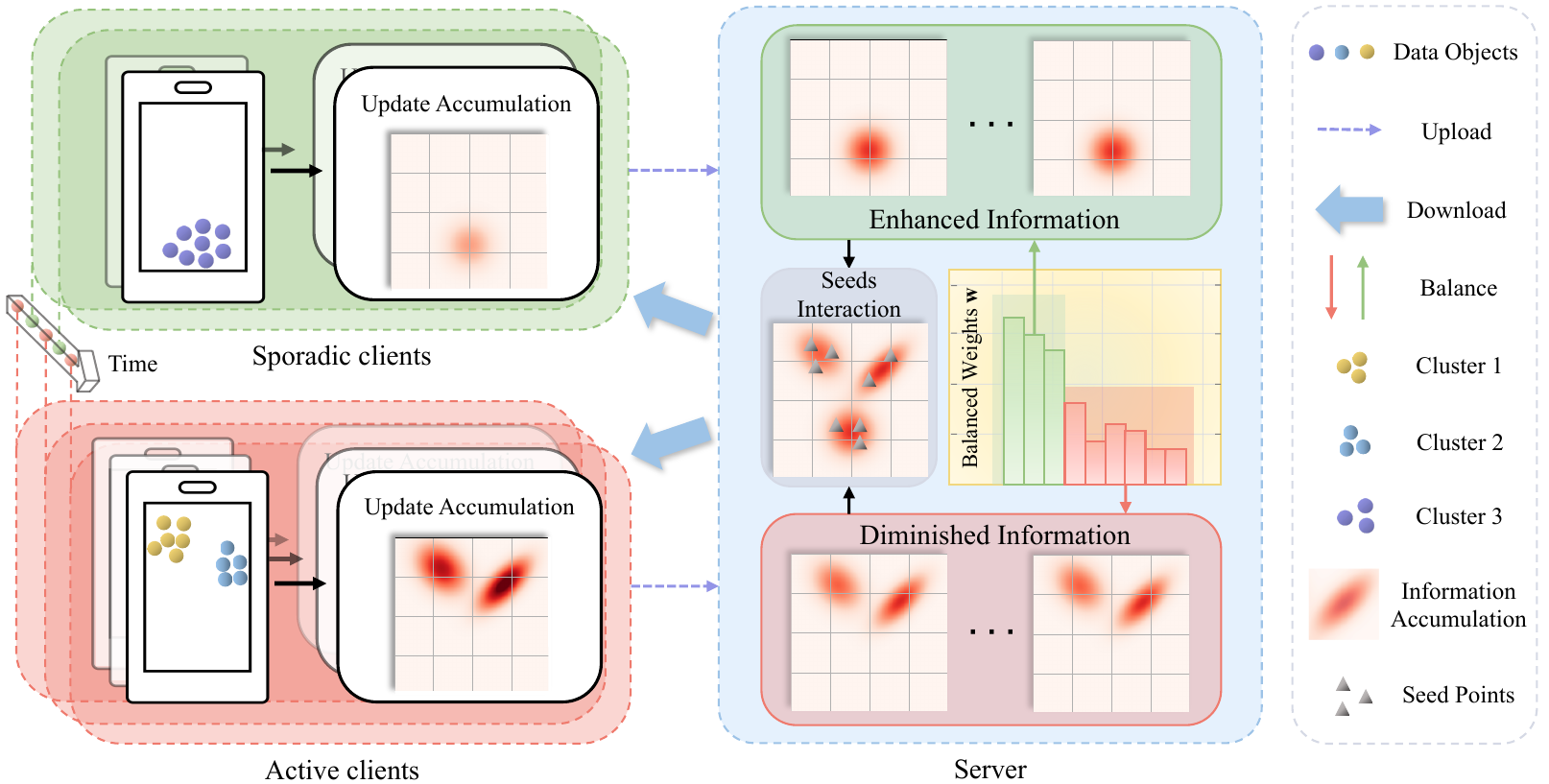}
\caption{Overview of the proposed AFCL framework. Initialized seed points accumulate update intensity from different clients independently, then the server balances the update information to facilitate appropriate seeds interaction for fusing the clients' distributions. The heat map represents the intensity of update information of seeds accumulated from asynchronous clients.}\label{fg:alg-all}
\end{figure*}

\subsection{Clustering with Unknown Cluster Number}
The more realistic unsupervised or weakly supervised learning has attracted much attention in recent years, especially for some significant application domains \cite{cheung2009local, wu2019unsupervised, EBDM-o-n, wang2023self}. Clustering is a key unsupervised learning technique, where the traditional clustering methods determine the optimal number of clusters $k^*$ manually \cite{Silhouettes1987, Elbow1953}. To realize automated $k^*$ selection, density-based clustering \cite{ester1996density, zhangcais2024, peng2025icassp} have been proposed to automatically determine $k^*$ at a ``knee point'' during their cluster exploration. Recently, more advanced learning-based approaches \cite{YMC-CCL, cheung2013categorical, cai2024robust} introduce a cooperative or competitive mechanism to excessive cluster centers. They simultaneously ensure the comprehensive representation of object distributions and make the elimination of redundant cluster centers learnable, thus achieving satisfactory clustering performance. Most recently, significance-based approaches \cite{hu2022significance, hu2025significance} have been proposed to rigorously judge the significance of cluster distributions under the current $k$. Nevertheless, all the above-mentioned solutions require detailed statistics of the entire dataset, which hamper their applicability in FC.

\section{Proposed Method}

\begin{table}[t]
    \centering
    \caption{Summary of notations.}
    \resizebox{0.93\linewidth}{!}{
    \begin{tabularx}{0.98\linewidth}{l|X}
    \toprule
        Notations & Explanations \\ 
        \midrule
        $\mathbf{X}$ & Global dataset\\
        $\mathbf{X}^{\{g\}}$ & Dataset of $g$-th client \\
        $\mathbf{M}^{\{g\}}$ & Seed points of $g$-th client\\
        $\mathbf{m}_l$ & $l$-th global seed point\\
        $\mathbf{Q}^{\{g\}}$ & Object-cluster affiliation matrix corresponding to $g$-th client \\
        $\mathbf{B}^{\{g\}}$ & Cluster center set of $g$-th client\\
        $\mathbf{R}^{\{g\}}_l$ & Update intensity of $l$-th seed on $g$-th client \\
        $k^*$ &  The `true' global cluster number of $\mathbf{X}$\\
        \bottomrule
        \end{tabularx}
    }
    \label{tb:notations}
\end{table}

In this section, we first define the asynchronous FC problem, and then present the proposed AFCL algorithm composed of two key technical components: 1) CSUA: Client-Side Update Accumulation, and 2) SSSI: Server-Side Seeds Interaction. Frequently used notations are summarized in Table~\ref{tb:notations}, and the overview of AFCL is shown in Fig.~\ref{fg:alg-all}.

\subsection{Problem Formulation}

Assuming a federated network with $p$ clients dividing the entire dataset $\mathbf{X}$ into the corresponding subsets $\{\mathbf{X}^{\{1\}}, \mathbf{X}^{\{2\}}, ..., \mathbf{X}^{\{g\}}, ..., \mathbf{X}^{\{p\}}\}$, where the subset of $g$-th client $\mathbf{X}^{\{g\}}$ has $n^{\{g\}}$ objects $\{\mathbf{x}^{\{g\}}_1, \mathbf{x}^{\{g\}}_2, ..., \mathbf{x}^{\{g\}}_{n^{\{g\}}}\}$ with $\sum^p_{g=1} n^{\{g\}} = n$, and each object $\mathbf{x}^{\{g\}}_{i}=[x^{\{g\}}_{i,1},x^{\{g\}}_{i,2},..., x^{\{g\}}_{i,d}]^\top$ is a $d$-dimensional vector. 
 We use a matrix $\mathbf{Q} \in \mathbb{R}^{n\times k}$ to indicate the object-cluster affiliation, and the conventional clustering objective is to minimize the overall intra-cluster dissimilarity between the objects and the cluster center (also called seed point or seed interchangeably hereinafter), which can be written as:
\begin{equation}\label{eq:obj1}
    Z(\mathbf{Q}, \mathbf{M}) = \sum^n_{i=1}\sum^k_{l=1} q_{i,l}\Phi(\mathbf{x}_i,\mathbf{m}_l), 
\end{equation}
where $\Phi(\mathbf{x}_i,\mathbf{m}_l)$ denotes the Euclidean distance between $i$-th object $\mathbf{x}_i$ and $l$-th seed $\mathbf{m}_l$. All the $k$ seeds can be organized as a matrix $\mathbf{M} \in \mathbb{R}^{d\times k}$ and $q_{i,l}$ is the $(i,l)$-th entry of $\mathbf{Q}$ satisfying ${\sum_{l=1}^{ k}q_{i,l}} = 1$ and $q_{i,l} \in \{0, 1\}$. For federated clustering, each $g$-th client can perform the above clustering locally on $\mathbf{X}^{\{g\}}$, and the ultimate goal is to minimize the objective function at the server with the entire dataset $\mathbf{X}$ and the consensus seed points $\mathbf{M}$ learned across all the clients.

\subsection{CSUA: Client-Side Update Accumulation}\label{sec:client}

To protect the privacy of clients while transmitting their distribution information to the server for global clustering, some intermediate quantities, e.g., update intensity of seeds, clustering center, and intra-cluster dissimilarity will be extracted by performing local clustering on each client and then uploaded to the server for seeds interaction.

For each object $\mathbf{x}^{\{g\}}_i$ in $g$-th client, its belonging cluster is determined by: 
\begin{equation}\label{eq:indicator}
    q_{i,l} = \begin{cases} 1, & \text{if}\ l = \mathop{\arg\min}\limits_r \gamma_r \parallel \mathbf{x}^{\{g\}}_i - \mathbf{m}^{\{g\}}_r\parallel ^2\\0, & {\rm otherwise}, 
    \end{cases}
\end{equation}
where $\mathbf{m}^{\{g\}}_r$ is the $r$-th seed in $\mathbf{M}^{\{g\}}$, and $\gamma^{\{g\}}_r$ is the weight of $\mathbf{m}^{\{g\}}_r$ computed by:
\begin{equation}\label{eq:gamma}
    \gamma^{\{g\}}_r = \frac{s^{\{g\}}_r}{\sum_{l=1}^{ k}s^{\{g\}}_l}.
\end{equation}
Here, $s^{\{g\}}_r$ denotes the winning time of $\mathbf{m}^{\{g\}}_r$ within a single iteration, which is updated by:
\begin{equation}\label{eq:s_r}
    s^{\{g\}}_c = s^{\{g\}}_c + 1,
\end{equation}
for each $q_{i,c}=1$, $i\in\{1,2,...,n^{\{g\}}\}$. 

To facilitate the interaction of seeds across clients, we also compute the update intensity of seeds locally, but suspend the update until they are uploaded to the server. For the winner seed $\mathbf{m}^{\{g\}}_c$, its update intensity $\mathbf{r}^{\{g\}}_{c,i}$ contributed by the object $\mathbf{x}^{\{g\}}_i$ is computed by:
\begin{equation}\label{eq:R_c_i}
    \mathbf{r}^{\{g\}}_{c,i} = \eta(\mathbf{x}^{\{g\}}_i - \mathbf{m}^{\{g\}}_c),
\end{equation}
where $\mathbf{r}^{\{g\}}_{c,i} \in \mathbf{R}^{\{g\}}_c$ and $\eta$ is the learning rate. By computing the update intensity of $k$ seeds provided by all the $n^{\{g\}}$ objects, we obtain $R^{\{g\}}= \{ \mathbf{R}^{\{g\}}_1, \mathbf{R}^{\{g\}}_2, ..., \mathbf{R}^{\{g\}}_k\}$ to upload to the server. 

\begin{remark}
\label{rm1}
\textbf{Privacy Protection w.r.t. $R^{\{g\}}$:} AFCL uploads $R^{\{g\}}$s to the server to facilitate the interaction among seeds for the fusion of clients' information. Since the update intensity of a $r$-th seed provided by each of the objects treating it as a winner (i.e., $\mathbf{R}^{\{g\}}_r$) will be uploaded, the risk of objects recovery and privacy leakage will be increased. Therefore, existing privacy-preserving techniques such as holomorphic encryption \cite{acar2018survey} and differential privacy \cite{wei2020federated, li2023differentially} can be incorporated in some scenarios with high privacy protection requirements. Specifically, these techniques can be utilized to perturb the update intensity provided by each object in $R^{\{g\}}$ while ensuring that the radius of cooperative seeds selection in Eq.~\eqref{eq:r} and the overall update of seeds in Eq.~\eqref{eq:update} unchanged. Note that this paper focuses on more robust FC, rather than improving its privacy protection level.
\end{remark}

To judge convergence at the server, we also compute the centers $\mathbf{B}^{\{g\}} = \{\mathbf{b}^{\{g\}}_1, \mathbf{b}^{\{g\}}_2, ...,\mathbf{b}^{\{g\}}_k \}$ of the $k$ clusters partitioned by the seeds by:
\begin{equation} \label{eq:b_l}
    \mathbf{b}^{\{g\}}_r = \frac{1}{o_r^{\{g\}}}\sum_{i=1}^{n^{\{g\}}}q_{i,r}\mathbf{x}^{\{g\}}_i,
\end{equation}
where $o_r^{\{g\}}$ is the number of objects in the $r$-th cluster corresponding to $\mathbf{m}^{\{g\}}_r$. Based on $\mathbf{B}^{\{g\}}$, the contribution of the $r$-th cluster to the global objective $Z$ can be computed by:
\begin{equation} \label{eq:z_r}
    z^{\{g\}}_r = \sum_{i=1}^{n^{\{g\}}}q_{i,r}\parallel\mathbf{b}^{\{g\}}_r - \mathbf{x}^{\{g\}}_i\parallel^2,
\end{equation}
and the contributions from all the seeds in $g$-th client can be collectively denoted as $\mathbf{z}^{\{g\}}=[z^{\{g\}}_1,z^{\{g\}}_2,...,z^{\{g\}}_k]^\top$.
\begin{remark}
    \textbf{Necessity of Intermediate $\mathbf{B}^{\{g\}}$ and $\mathbf{z}^{\{g\}}$:} Since the trajectories of seeds $\mathbf{M}$ is usually complex as shown in Fig.~\ref{vi}(d), directly calculating the objective function based on seeds may lead to constant fluctuations in the objective function value. Therefore, intermediate values $\mathbf{B}^{\{g\}}$ and $\mathbf{z}^{\{g\}}$ computed locally on each client are relatively stable and can timely reflect the goodness of current seeds in terms of each client, which are helpful to obtain a smooth and more approximate overall objective function value $Z$.
\end{remark}

\subsection{SSSI: Server-Side Seeds Interaction}

During the learning, communication frequencies of each client should be recorded as $\mathbf{\Theta} = [\theta^{\{1\}}, \theta^{\{2\}}, .., \theta^{\{p\}}]^\top$. Accordingly, weights $\mathbf{w} = [w^{\{1\}}, w^{\{2\}}, ..., w^{\{p\}}]^\top$ for balancing the seeds update bias caused by the asynchronous communication should be maintained. For $g$-th participating client, its balance weight is computed by:
\begin{equation}\label{eq:com_w}
w^{\{g\}} = \frac{\xi}{\xi + \frac{\theta^{\{g\}}}{\sum^p_{j=1}\theta^{\{j\}} }},
\end{equation}
where $\xi$ is a hyper-parameter controlling the sensitivity of balance weight w.r.t. the communication frequencies $\mathbf{\Theta}$. Intuitively, a larger $\xi$ makes the balance weight less sensitive to the frequency, and a client with a higher frequency will have a lower weight to weaken its contribution.



Upon receiving the uploading from clients, the server initially aggregates the cluster centers and their objective contributions with the balance weights by:
\begin{equation}\label{eq:agg-br}
    \mathbf{b}_r = \sum_{j=1}^{\Bar{p}}\frac{w^{\{j\}}o_r^{\{j\}} \mathbf{b}^{\{j\}}_r}{\sum_{j=1}^{\Bar{p}}o_r^{\{j\}}}, \  \ z_r = \sum_{j=1}^{\Bar{p}}\frac{w^{\{j\}}o_r^{\{j\}} z^{\{j\}}_r}{\sum_{j=1}^{\Bar{p}}o_r^{\{j\}}},
\end{equation} 
where $\Bar{p}$ denotes the number of participating clients. $\mathbf{B} = \{\mathbf{b}_1, \mathbf{b}_2, ...,\mathbf{b}_k\}$ and $\mathbf{z} = [z_1, z_2, ...,z_k]^\top$ represent the aggregated centers and contributions to the objective $Z$ from the server perspective. Accordingly, an approximation of the objective on the server can be derived from Eq.~\eqref{eq:obj1} as:
\begin{equation}\label{eq:convergence}
    Z(\mathbf{B},\mathbf{z}) = \frac{1}{k}\sum_{l=1}^{{k}}\mathop{\max}\limits_{r\neq l}\left(\frac{z_l + z_r}{\parallel \mathbf{b}_l - \mathbf{b}_r\parallel^2}\right),
\end{equation}
which is in the form of the DBI index \cite{ros2023pdbi} that simultaneously reflects cluster compactness (numerator) and clusters' dispersion (denominator). 


To ensure that the seeds learned from data can consensually minimize $Z$, a cooperative set:
\begin{equation}\label{eq:c}
    C_r = \{\mathbf{m}_l\mid \parallel \mathbf{m}_r - \mathbf{m}_l\parallel^2 \leq \parallel \mathbf{m}_r - \mathbf{x}^{\{g\}}_i\parallel^2\}
\end{equation}
is identified for each seed $\mathbf{m}_r$, and we let all the seeds 
in $C_r$ collectively receive the updates from the samples 
as:
\begin{equation}\label{eq:wu}
    \mathbf{m}_{u} = \mathbf{m}_{u} + \eta(\mathbf{x}^{\{g\}}_i - \mathbf{m}_u),
\end{equation}
where $\mathbf{m}_u \in C_r$. However, since original samples are unavailable at the server due to privacy constraints, 
we compute Eqs.~\eqref{eq:c} and~\eqref{eq:wu} alternatively based on the update intensity $\mathbf{r}^{\{g\}}_{r,i}$ that has been uploaded to the server as:
\begin{equation}\label{eq:r}
    C_r = \{\mathbf{m}_l\mid \parallel \mathbf{m}_r - \mathbf{m}_l\parallel^2 \leq \parallel \frac{w^{\{g\}} \mathbf{r}^{\{g\}}_{r,i}}{\eta}\parallel^2\}
\end{equation}
and 
\begin{equation}\label{eq:update}
    \mathbf{m}_{u} = \mathbf{m}_{u} + w^{\{g\}}\mathbf{r}^{\{g\}}_{r,i} + w^{\{g\}}\eta(\mathbf{m}_r - \mathbf{m}_u)
\end{equation}
respectively, where $w^{\{g\}}$ is used to mitigate the bias of seeds update caused by the asynchronous uploading. Intuitively, a smaller $w^{\{g\}}$ corresponds to a client that uploads frequently. Thus its neighbor identifying radius $\parallel w^{\{g\}} \mathbf{r}^{\{g\}}_{r,i}/\eta\parallel^2$ in Eq.~\eqref{eq:r} will be smaller to avoid over-update of the seeds in $C_r$. In Eq.~\eqref{eq:update}, $\mathbf{r}^{\{g\}}_{r,i}$ from Eq.~\eqref{eq:R_c_i} is already with the learning rate $\eta$, and thus its latter two terms are with the same coefficients $w^{\{g\}}\eta$, equivalent to update $\mathbf{m}_{u}$ by a small step towards the data object that yields the update intensity $\mathbf{r}^{\{g\}}_{r,i}$.


\begin{remark}
    \textbf{Interaction of Seeds:} According to Eqs.~\eqref{eq:r} and~\eqref{eq:update}, all seeds in $C_r$ will move closer to the cluster distribution represented by $\mathbf{m}_r$ according to the distribution information accumulated on different clients, which facilitates the distribution completion across different clients.
\end{remark}

\begin{algorithm}[t]
\caption{AFCL: Asynchronous Federated Clustering.} 
\label{alg:all}
\textbf{Input}: $\mathbf{X}^{\{1\}}, \mathbf{X}^{\{2\}}, ..., \mathbf{X}^{\{p\}}$, $k$, $\xi$, $\eta$.
\begin{algorithmic}[1] 
 \Statex \boxit{mypink}{4.49}
 \vspace{-4mm}
 \For{\textbf{all clients} in parallel} 
    \State Initialize $k$ seeds using $k$-means++;
    \State Transfer initialized seeds to the server;
\EndFor
 \vspace{0.4mm}
 \Statex \boxit{myblue}{6.86}
 \vspace{-4mm}
\State Initialize $k$ global seeds $\mathbf{M}$ using $k$-means++ according to the seeds received from all clients;
\Repeat
    \State Transfer global seeds $\mathbf{M}$ to participating clients;
     \vspace{0.4mm}
     \Statex \boxit{mypink}{5.49}
     \vspace{-4mm}
\For{\textbf{$g$-th participating client} in parallel}
    \State Update $\theta^{\{g\}} = \theta^{\{g\}} + 1$;
    \State  Compute $\mathbf{Q}^{\{g\}}, {R}^{\{g\}}, \mathbf{B}^{\{g\}}, \mathbf{z}^{\{g\}}$, by Eqs.~\eqref{eq:indicator}, 
    \Statex \ \ \ \ \ \ \ \ \ \ \ \ \eqref{eq:R_c_i}, \eqref{eq:b_l}, and \eqref{eq:z_r}, respectively;
    \State  Upload $\mathbf{Q}^{\{g\}}, {R}^{\{g\}}, \mathbf{B}^{\{g\}}, \mathbf{z}^{\{g\}}$ to the server;
    \EndFor 
    \vspace{0.4mm}
    \Statex \boxit{myblue}{2.1}
    \vspace{-4mm}
    \State Compute $\mathbf{M}$, $Z$ using Eqs.~\eqref{eq:update} and \eqref{eq:convergence};
\State Update $\mathbf{w}$ by Eq.~\eqref{eq:com_w};
\Until{Convergence}
\end{algorithmic}
\textbf{Output}: $\mathbf{M}$, $\mathbf{Q}$.
\end{algorithm}


\subsection{Overall AFCL Algorithm}\label{sct:overall}

In the AFCL algorithm, after the global initialization of the seed points and local update accumulation on all the clients, sufficient interaction among the clients through the server is iteratively performed until the convergence of $Z$. In this process (summarized as Algorithm~\ref{alg:all}), the client-side (indicated by pink color) mainly implements cluster distribution learning, and the server-side (indicated by blue color) is responsible for privacy-protected distribution information fusion. For each learning iteration of the AFCL algorithm, the time complexity is $\mathcal{O}(kn^{\{g\}}dp + n^{\{g\}}k^2d)$, which is efficient compared to the state-of-the-art FC methods. A more detailed analysis can be found in the Appendix.

\begin{links}
    \link{Appendix}{https://github.com/Yunfan-Zhang/AFCL}
\end{links}


\section{Experiments}

\subsection{Experimental Setup}

\textbf{Six experiments} have been conducted: (1) Visualization: Intuitive demonstration of the learning process of the proposed AFCL; (2) Convergence Evaluation: Objective function values at each learning iteration are recorded to illustrate the convergence and efficiency of AFCL; (3) Clustering Performance Evaluation: We compare AFCL with the conventional and state-of-the-art counterparts to demonstrate the superiority of AFCL; 
Due to space limitation, the remainder three experiments, i.e., significance test, ablation study, execution time evaluation, and more detailed experimental settings, are provided in the online appendix.

\begin{table}[t]
    \centering
    \caption{Statistics of experimental datasets.}
    \resizebox{1\linewidth}{!}{
    \begin{tabular}{c|cc|ccc}
    \toprule
        No. & Datasets & Abbrev. & $n$ & $d$ & $k^*$ \\ 
        \midrule
        1 & Synthetic Dataset 1 & SD1 & 2300 & 2 & 4 \\
        2 & Synthetic Dataset 2 & SD2 & 2900 & 2 & 5 \\
        3 & Seeds \cite{misc_seeds_236} & SE & 210 & 7 & 3 \\
        4 & Iris \cite{misc_iris_53} & IR & 150 & 4 & 3 \\
        5 & Avila \cite{misc_avila_459} & AL & 10430 & 10 & 12 \\ 
        6 & Abalone \cite{rad2019hybrid} & AB & 4177 & 7 & 29 \\ 
        7 & Breast Cancer \cite{misc_breast_cancer_wisconsin} & CC & 569 & 30 & 2 \\ 
        8 & Accent \cite{misc_speaker_accent_recognition_518} & AC & 329 & 12 & 6 \\
        9 & Segment \cite{misc_image_segmentation_50} & SG & 2100 & 19 & 7 \\
        10 & Live \cite{misc_facebook_live_sellers_in_thailand_488} & LI & 7051 & 9 & 2 \\
        11 & Parkinson \cite{misc_parkinsons_174}& PA & 197 & 22 & 2 \\ 
        12 & Audit \cite{misc_audit_data_475} & AU & 776 & 24 & 2\\ 
        13 & Transfusion \cite{misc_blood_transfusion_service_center_176}& TF & 748 & 4 & 2\\
        \bottomrule
    \end{tabular}
    }
    \label{tb:st}
\end{table}

\begin{figure}[t]
\centering
    \subfigure[Client 1]{
        \includegraphics[width=0.21\textwidth]{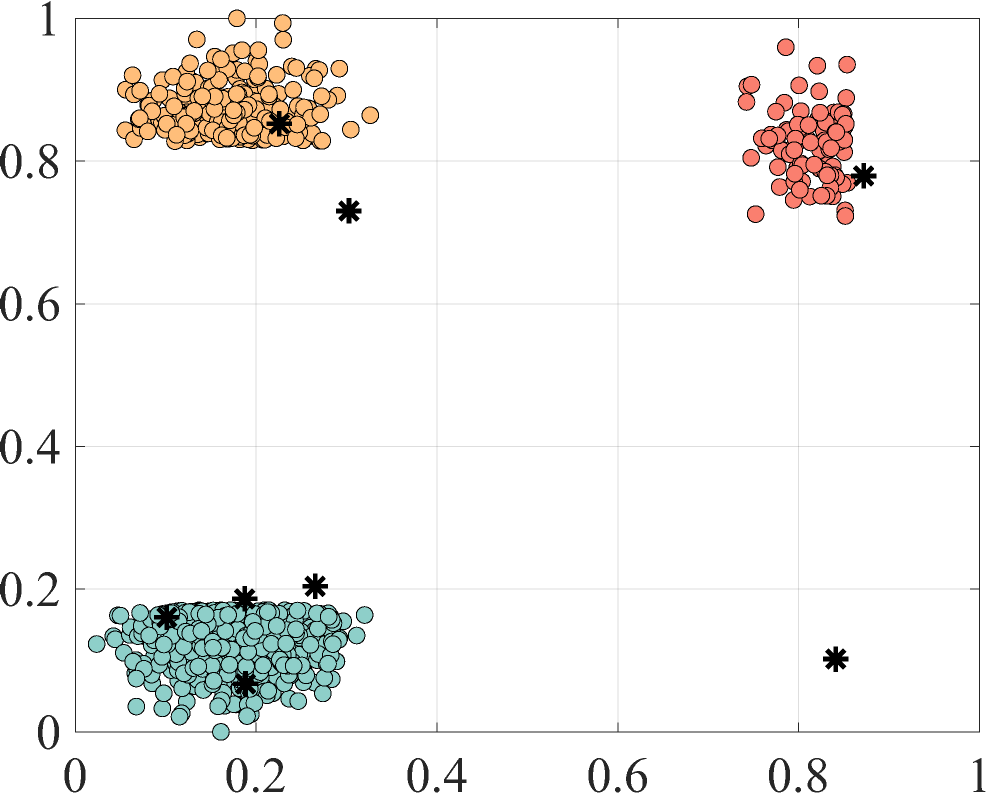}
        \label{subfig:v1}
    }
    \subfigure[Client 2]{
        \includegraphics[width=0.21\textwidth]{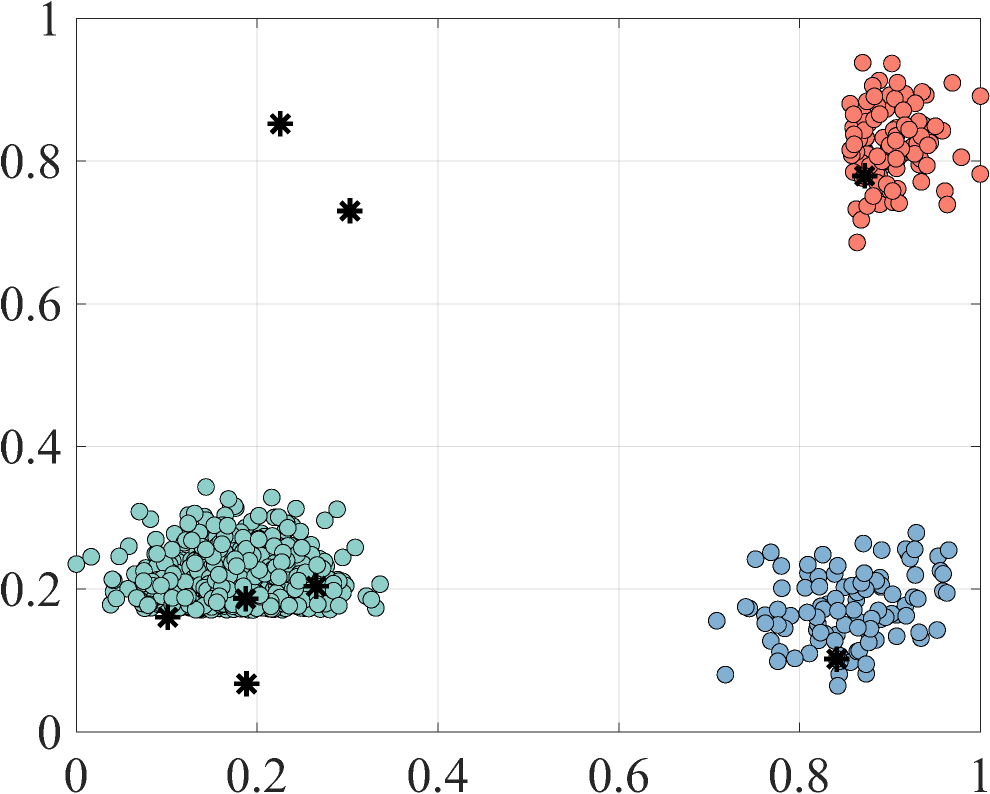}
        \label{subfig:v2}
    }
        \subfigure[Client 3]{
        \includegraphics[width=0.21\textwidth]{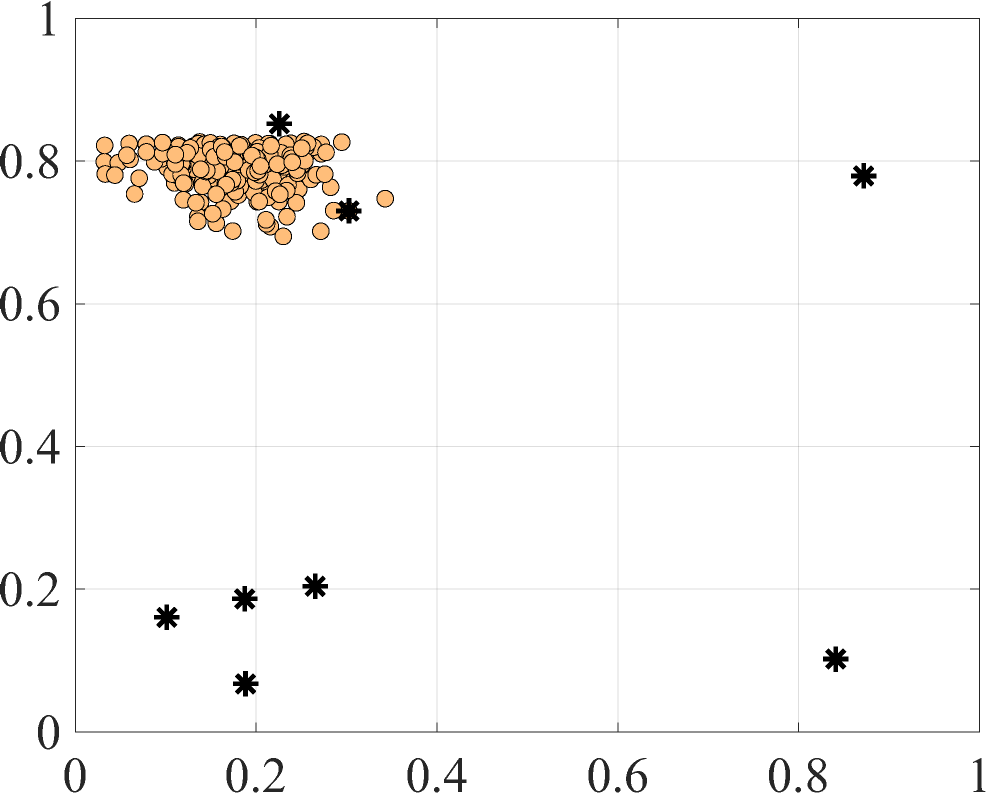}
        \label{subfig:v3}
    }
    \subfigure[Process on Server]{
        \includegraphics[width=0.21\textwidth]{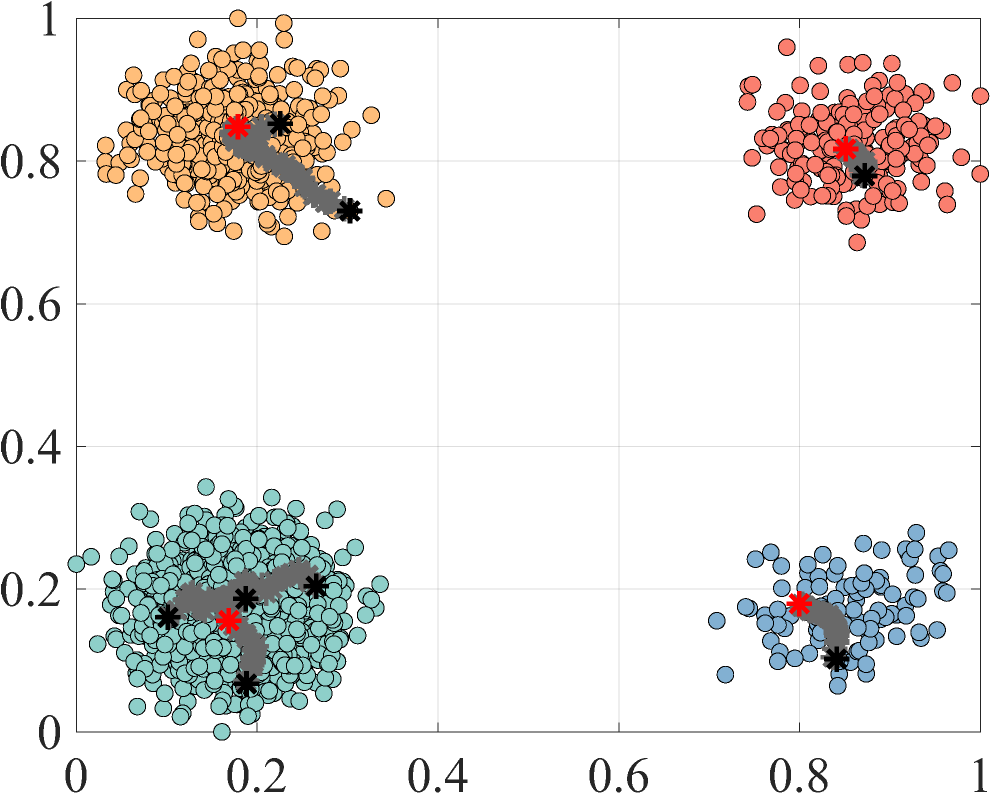}
        \label{subfig:v4}
    }
\caption{Seed points and their trajectories on the server during the learning of AFCL. Black and red dots indicate the initial and final positions of the seed points, respectively.}
\label{vi}
\end{figure}

\textbf{Six counterparts} are compared: DK++ \cite{bahmani2012scalable} is a conventional distributed learning approach that conforms to the settings of FL. Five state-of-the-art methods, i.e., the iterative learning approaches FFCM-avg1, FFCM-avg2 \cite{FFCM}, and FedSC \cite{qiao2024federated}, and the one-shot learning approaches $k$-FED \cite{k-fed} and Fed-SC \cite{xie2023fed}, are compared. All their hyper-parameters (if any) are set according to the corresponding source papers.

\textbf{13 datasets}, including two Gaussian spherical synthetic datasets, and 11 public real datasets collected from the UCI machine learning repository \cite{asuncion2007uci}, are utilized for the experiments, and their statistics are shown in Table~\ref{tb:st}. All the real datasets are pre-processed by omitting the objects with missing values and normalized using min-max scaling. 

\textbf{Three validity indices} including the Silhouette Coefficient index (SC) \cite{Silhouettes1987}, Calinski-Harabasz index (CH)~\cite{CH1974}, and Bonferroni–Dunn (BD) test \cite{demvsar2006statistical} are chosen for performance evaluation. Values of SC and CH are in the intervals [-1,1] and (0,$+\infty$), respectively, with a higher value indicating a better clustering performance. These two internal indices are insensitive to the number of clusters, thus facilitating a fair comparison of AFCL that learns its own $k$ and the counterparts with a pre-set $k^*$ as shown in Table~\ref{tb:st}. BD test is conducted on the performance ranks of the counterparts to validate if AFCL performs significantly better. 
 
\subsection{Visualization}

\begin{figure}[!t]
\centering
    \subfigure[SD1 Dataset]{
        \includegraphics[width=0.22\textwidth]{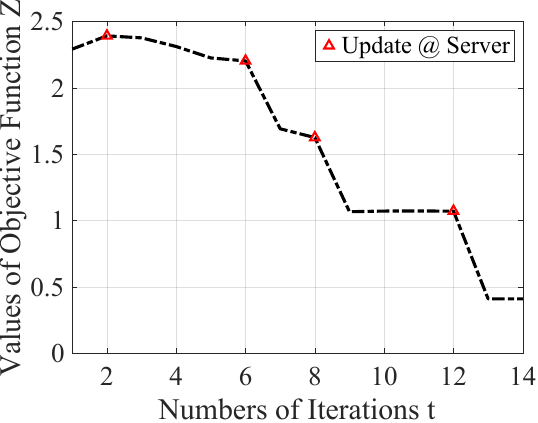}
        \label{subfig:con1}
    }
    \subfigure[IR Dataset]{
        \includegraphics[width=0.22\textwidth]{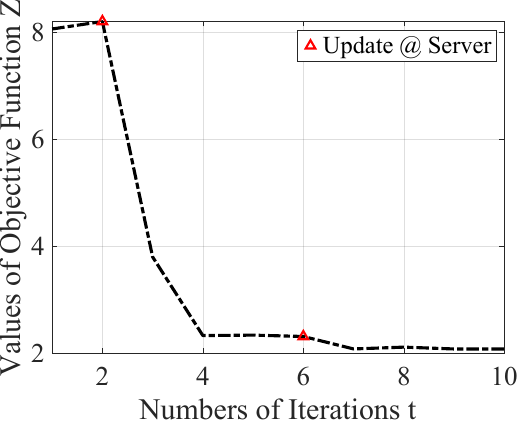}
        \label{subfig:con2}
    }
        \subfigure[SE Dataset]{
        \includegraphics[width=0.22\textwidth]{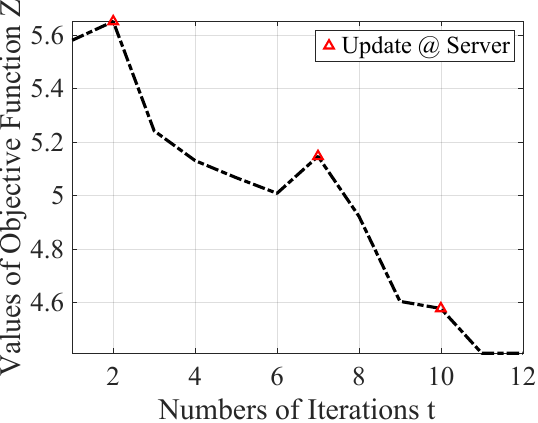}
        \label{subfig:con3}
    }
        \subfigure[AL Dataset]{
        \includegraphics[width=0.22\textwidth]{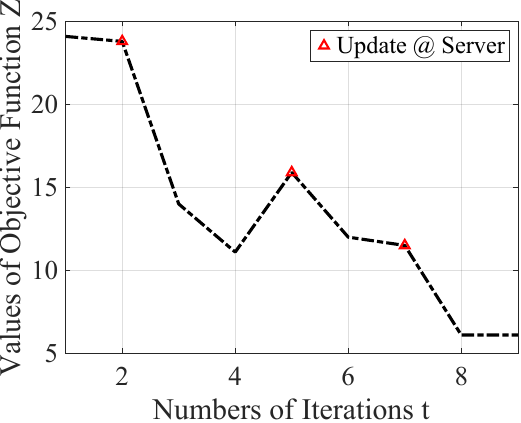}
        \label{subfig:con4}
    }
\caption{Values of the AFCL objective function on (a) SD1, (b) IR, (c) SE, and (d) AL datasets. Red triangles mark the iterations that the server update starts.}
\label{fg:obj}
\end{figure}

\begin{table*}[!t]
\centering
\caption{Clustering performance evaluated by SC on all the 13 datasets.}
\resizebox{1.86\columnwidth}{!}{ 
    \begin{tabular}{c|ccccccc}
    \toprule
    Dataset & DK++ & $k$-FED & FFCM-avg1 & FFCM-avg2  & Fed-SC & FedSC & AFCL \\ 
    \midrule

        SD1 \ & 0.5986±0.18 & 0.8494±0.00 & 0.5063±0.02 & 0.5036±0.02 & 0.8261±0.09 & \underline{0.8539±0.00} & \textbf{0.9714±0.00} \\ 
       SD2 \ & 0.6127±0.11 & \underline{0.7699±0.00} & 0.4773±0.03 & 0.4679±0.03 & 0.6005±0.14 & 0.7477±0.00 & \textbf{0.8571±0.00} \\ 
        SE \ & 0.3229±0.00 & 0.3754±0.04 & \underline{0.4323±0.05} & \underline{0.4323±0.05} & 0.3619±0.00 & 0.3774±0.00 & \textbf{0.5033±0.09} \\ 
        IR \ & 0.4955±0.01 & 0.4818±0.02 & 0.5672±0.02 & \underline{0.6119±0.21} & 0.5384±0.04 & 0.5719±0.00 & \textbf{0.6386±0.06} \\ 
        AL \ & 0.2096±0.02 & 0.0979±0.03 & 0.2721±0.09 & 0.2761±0.07 & \underline{0.4422±0.05} & 0.3187±0.00 & \textbf{0.6138±0.38} \\ 
        AB \ & 0.2314±0.02 & 0.1853±0.01 & 0.3664±0.34 & \underline{0.4863±0.26} & 0.3009±0.03 & 0.3865±0.06 & \textbf{0.5005±0.11} \\ 
        CC\  & 0.3778±0.00 & \underline{0.3809±0.02} & 0.2873±0.05 & 0.3051±0.04 & 0.3672±0.04 & 0.3747±0.00 & \textbf{0.5916±0.04} \\ 
        AC\  & 0.1831±0.02 & 0.0992±0.01 & \underline{0.2656±0.13} & 0.2358±0.13 & 0.1376±0.02 & 0.1922±0.00 & \textbf{0.4851±0.26} \\ 
        SG \ & 0.3197±0.02 & 0.3117±0.00 & 0.3819±0.11 & \underline{0.3865±0.10} & 0.2677±0.04 & 0.2935±0.00 & \textbf{0.6086±0.12} \\ 
        LI \ & 0.8241±0.00 & \underline{0.8256±0.01} & 0.4805±0.16 & 0.4239±0.23 & 0.7980±0.10 & 0.8252±0.00 & \textbf{0.8967±0.01} \\ 
        PA \ & 0.2763±0.00 & 0.4517±0.03 & 0.4419±0.15 & 0.4419±0.15 & 0.2443±0.00 & \textbf{0.5138±0.00} & \underline{0.4903±0.11} \\ 
        AU \ & 0.4046±0.05 & 0.3601±0.04 & 0.1440±0.08 & 0.2239±0.02 & 0.3411±0.07 & \underline{0.4296±0.00} & \textbf{0.5191±0.07} \\ 
        TF \ & 0.4935±0.00 & 0.5269±0.01 & \underline{0.6402±0.16} & \underline{0.6402±0.16} & 0.5172±0.25 & 0.5390±0.00 & \textbf{0.6813±0.03} \\ \midrule
        Ave. Rank & 4.6154 & 4.2308 & 4.4231 & 4.1923 & 5.4615 & \underline{4.0000} &  \textbf{1.0769} \ \\ 
        \bottomrule
    \end{tabular}
    }
    \label{tb:sc}
\end{table*}

\begin{table*}[t]
    \centering
    \caption{Clustering performance evaluated by CH on all the 13 datasets.}
    \resizebox{1.86\columnwidth}{!}{ 
    \begin{tabular}{c|R{1.8cm}R{1.8cm}R{1.8cm}R{1.8cm}R{1.8cm}R{1.8cm}R{1.8cm}}
    \toprule
        \ Dataset\  &\ \ \ \ DK++& $k$-FED& FFCM-avg1 & FFCM-avg2 & Fed-SC & FedSC & AFCL \\
        \midrule
        \ SD1 \ & \ 13933.3423 & 18933.7121 & 3136.2342 & 3140.0656 & 18901.0798 & \underline{18958.6529} & \textbf{19482.8610} \ \\ 
        \ SD2 \ & \ 13187.1700 & \underline{16936.0195} & 1462.2227 & 1474.5426 & 1575.4188 & 16913.4103 & \textbf{17200.3823} \ \\ 
        \ SE \ & \ 192.6124 & \textbf{251.1952} & 83.3635 & 83.3635 & 193.6615 & 206.0449 & \underline{230.9555} \ \\ 
        \ IR\  & \ \textbf{315.2151} & 232.9987 & 65.6174 & 66.3026 & 211.5583 & 232.9386 & \underline{310.7035} \ \\ 
        \ AL\  & \ 1524.2576 & 1559.4321 & 1607.4565 & 1809.4353 & \underline{2260.6359} & 1524.2576 & \textbf{4880.6220} \ \\ 
        \ AB\  & \ 3756.1887 & 3791.0247 & 3821.2626 & \underline{3890.4190} & 3072.4473 & 3244.3141 & \textbf{5906.3378} \ \\ 
        \ CC\  & \ 202.4573 & \textbf{290.0258} & 104.0924 & 107.4325 & 222.6533 & 231.4785 & \underline{231.5775} \ \\ 
        \ AC\  & \ \underline{112.3225} & 80.9425 & 82.4193 & 77.3857 & 73.2801 & 99.9817 & \textbf{140.7156} \ \\ 
        \ SG\  & \ 973.7952 & 1121.9421 & 765.2036 & \underline{1278.2297} & 781.12178 & 1057.1972 & \textbf{1541.6809} \ \\ 
        \ LI\  & \ 5013.7432 & \underline{5526.4145} & 1050.4084 & 806.5331 & 4075.7079 & 3608.0879 & \textbf{6090.9982} \ \\ 
        \ PA\  & \ 40.2442 & 68.3886 & \underline{79.5121} & \underline{79.5121} & 56.2801 & \textbf{83.3207} & 78.1141\  \\ 
        \ AU\  & \ 304.9842 & 251.1567 & 65.7637 & 107.1965 & 201.6834 & \textbf{433.3965} & \underline{306.7529} \ \\ 
        \ TF\  & \ 393.1126 & \underline{540.1820} & 436.6372 & 436.6372 & 420.2884 & 518.4788 & \textbf{651.6017} \ \\ \midrule
        \ Ave. Rank\  & 4.5000 & \underline{3.0769} & 5.5769 & 4.8846 & 5.0000 & 3.4231 & \textbf{1.5385}\ \  \\ \bottomrule
    \end{tabular}
    }
    \label{tb:ch}
\end{table*}

To intuitively validate the effectiveness of AFCL, we split SD1 into three subsets for creating three extremely non-IID clients as shown in Fig.~\ref{subfig:v1} -~\subref{subfig:v3}. Fig.~\ref{subfig:v4} shows the global data distributions and update trajectories of the seeds on the server. It can be observed that even though the three clients have completely non-overlapping distributions, AFCL can still appropriately learn a set of seeds to represent the global cluster distributions. The trajectories also demonstrate that the seeds updating mechanism of AFCL can effectively facilitate interaction among the seeds with imbalanced update information uploaded by the asynchronous clients. After a certain number of learning iterations, redundant seeds are homogenized, i.e., overlapped at the center of several prominent clusters. This intuitively demonstrates the autonomous cluster number selection ability of AFCL.

To demonstrate the convergence efficiency of AFCL, we plot its objective function values on four datasets in Fig.~\ref{fg:obj}. It can be observed that AFCL converges quickly with around 10 iterations in most cases. Moreover, the objective function always experiences a steep decline after the server updates, confirming that the designed seeds interaction mechanism is highly effective. It is also noteworthy that, since only limited statistics are permitted to be communicated between clients and server, a strict gradient descent cannot be guaranteed and thus the convergence curve in Fig.~\ref{fg:obj} is not monotonically decreasing. Such an effect is rational for FC because the clustering objective can be viewed as heterogeneous at different clients and server.

\subsection{Clustering Performance Evaluation}

The clustering performance of AFCL and the existing FC approaches are compared under the non-IID and asynchronous scenarios. For each dataset, we implement clustering by using each compared method by 20 times and report the average performance. For each of the 20 trails, we first implement $k$-means with $k=5$, to divide the whole dataset into five clients to simulate the extreme non-overlapping distributions of non-IID clients. Then to simulate the asynchronous participation of clients, we randomly set each client with a different participation probability for controlling their upload in each iteration during the learning. As AFCL does not require $k^*$, the initial number of seed points $k$ is randomly selected from the range $[k^*, 2k^*]$. The clustering performance in terms of SC and CH obtained under the above settings is shown in Tables~\ref{tb:sc} and~\ref{tb:ch}. The best and the second-best results are highlighted using boldface and underline, respectively. The `Ave. Rank' rows report the average rank of different approaches across all datasets.

It can be observed that AFCL outperforms all its counterparts in general, indicating its superiority in asynchronous FC. Specifically, AFCL performs the best on almost all datasets w.r.t. the SC index, except on the PA dataset where AFCL still performs the second best. This is because AFCL can effectively minimize the intra-cluster dissimilarity and maximize the inter-cluster dispersion to search for the global optimal seeds. For the CH index, although AFCL performs the second-best on some datasets, i.e., SE, IR, CC, and AU, the best counterparts differ on these datasets while AFCL remains competitive in most cases. More specifically, for the cases where AFCL does not perform the best, the performance gap between AFCL and the best-performing counterpart is usually tiny, which demonstrates the effectiveness and robustness of AFCL on different datasets.

\section{Concluding Remarks}

This paper proposes a new FC approach called AFCL for mining global cluster distributions upon heterogeneous data distributions of asynchronously communicated clients. It advances FC to a more challenging but realistic scenario, i.e., clients can participate in the client-to-server uploading asynchronously, and all the clients and server can be extremely non-IID without knowing the `true' number of clusters. AFCL achieves this by adopting a client-to-seed information fusion framework, which lets the seed points cooperate on the server to complete the non-IID distributions of clients and automatically learns to eliminate redundant seeds as well. A balance mechanism is also designed to relieve the non-uniform of the update information uploaded by the asynchronously participated clients. As a result, AFCL can effectively outline the global cluster distributions upon the seeds learned by the aggregated update intensity received from clients, even if the communication is extremely asynchronous and the distributions of clients are completely divergent. Comprehensive experiments have illustrated the efficacy of AFCL. Despite the superiority of AFCL, there are still some noteworthy potential limitations. That is, we assume FC on pure numerical data and the number of clients is relatively small. The next promising avenue would be the FC of datasets comprising both numerical and categorical attributes distributed on a large number of clients. 

\section{Acknowledgments}
This work was supported in part by the National Natural Science Foundation of China (NSFC) under grants: 62476063, 62102097, 62376233, and 62306181, the NSFC/Research Grants Council (RGC) Joint Research Scheme under grant: N\_HKBU214/21, the Natural Science Foundation of Guangdong Province under grants: 2024A1313010039, 2024A1515010163, and 2023A1515012855, the Natural Science Foundation of Fujian Province under grant: 2024J09001, the General Research Fund of RGC under grants: 12201321, 12202622, and 12201323, the RGC Senior Research Fellow Scheme under grant: SRFS2324-2S02, the Shenzhen Science and Technology Program under grant: RCBS20231211090659101, the National Key Laboratory of Radar Signal Processing under grant: JKW202403, and the Xiaomi Young Talents Program.

\bibliography{FedCCL_AAAI}

\end{document}